\documentclass[11pt]{article} 
\usepackage{rldmsubmit,palatino}
\usepackage{graphicx}

\usepackage{subfigure}

\usepackage{amsmath,amsfonts,bm}

\def\eqref#1{equation~\ref{#1}}

\def\1{\bm{1}}

\DeclareMathAlphabet{\mathsfit}{\encodingdefault}{\sfdefault}{m}{sl}
\SetMathAlphabet{\mathsfit}{bold}{\encodingdefault}{\sfdefault}{bx}{n}

\DeclareMathOperator*{\argmax}{arg\,max}

\usepackage{titlesec}

\usepackage{enumitem}
\usepackage[numbers]{natbib}

\title{DynoPlan: Combining Motion Planning and Deep Neural Network based Controllers for Safe HRL}

\author{
Daniel Angelov \\
School of Informatics\\
University of Edinburgh\\
\texttt{d.angelov@ed.ac.uk} \\
\And
Yordan Hristov \\
School of Informatics\\
University of Edinburgh\\
\texttt{y.hristov@ed.ac.uk} \\
\And
Subramanian Ramamoorthy \\
School of Informatics\\
University of Edinburgh\\
\texttt{s.ramamoorthy@ed.ac.uk}
}

\begin{document}

\maketitle

\begin{abstract}
Many realistic robotics tasks are best solved compositionally, through control architectures that sequentially invoke primitives
and achieve error correction through the use of loops and conditionals taking the system back to alternative earlier states. 
Recent end-to-end approaches to task learning attempt to directly learn a single controller that solves an entire task, but this
has been difficult for complex control tasks that would have otherwise required a diversity of local primitive moves, and the resulting
solutions are also not easy to inspect for plan monitoring purposes. In this work, we aim to bridge the gap between hand designed and
learned controllers, by representing each as an option in a hybrid hierarchical Reinforcement Learning framework -
DynoPlan. We extend the options framework by adding a dynamics model and the use of a nearness-to-goal heuristic, derived from
demonstrations. This translates the optimization of a hierarchical policy controller to a problem of planning with a model
predictive controller. By unrolling the dynamics of each option and assessing the expected value of each future state, we
can create a simple switching controller for choosing the optimal policy within a constrained time horizon similarly to hill
climbing heuristic search. The individual dynamics model allows each option to iterate and be activated independently
of the specific underlying instantiation, thus allowing for a mix of motion planning and deep neural network based primitives. We
can assess the safety regions of the resulting hybrid controller by investigating the initiation sets of the different options,
and also by reasoning about the completeness and performance guarantees of the underpinning motion planners.
\end{abstract}


\acknowledgements{This research is supported by the Engineering and Physical Sciences Research Council (EPSRC), as part of the CDT in Robotics and Autonomous Systems at Heriot-Watt University and The University of Edinburgh. Grant reference EP/L016834/1., and by an Alan Turing Institute sponsored project on Safe AI for Surgical Assistance. }

\startmain 
\titlespacing\section{-5pt}{5pt plus 4pt minus 2pt}{-2pt plus 2pt minus 2pt}
\section{Introduction}

Open world tasks  often involve sequential plans. The individual steps in the sequence are usually quite independent from each other, hence can be solved through a number of different methods, such as motion planning approaches for reaching, grasping, picking and placing, or through the use of end-to-end neural network based controllers for a similar variety of tasks.
In many practical applications, we wish to combine such a diversity of controllers. This requires them to share a common domain representation.
For instance the problem of assembly can be represented as motion planning a mechanical part in proximity to an assembly and subsequently the use of a variety of wiggle policies to fit together the parts.
Alternatively, an end-to-end policy can be warm-started by using samples from the motion planner, which informs how to bring the two pieces together and the alignment sub-policy needed, as in \cite{thomas2018learning}. The resulting policy is {\textit{robust}} in the sense that the task of bringing together the assembly can be achieved from a large set of initial conditions and perturbations.

A hybrid hierarchical control strategy, in this sense, allows for different capabilities to be independently learned and composed into a task solution with multiple sequential steps. We propose a method that allows for these individual steps to consist of commonly used motion planning techniques as well as deep neural network based policies that are represented very differently from their sampling based motion planning counterparts. We rely on these controllers to have a dynamic model of the active part of their state space, and a sense of how close they are to completing the overall task. This allows the options based controller to predict the future using any of the available methods and then determine which one would bring the world state to one closest to achieving the desired solution - in the spirit of model based planning.

\begin{figure}[ht]
\centering
\subfigure[Gear Assembly]{\includegraphics[width=0.24\linewidth]{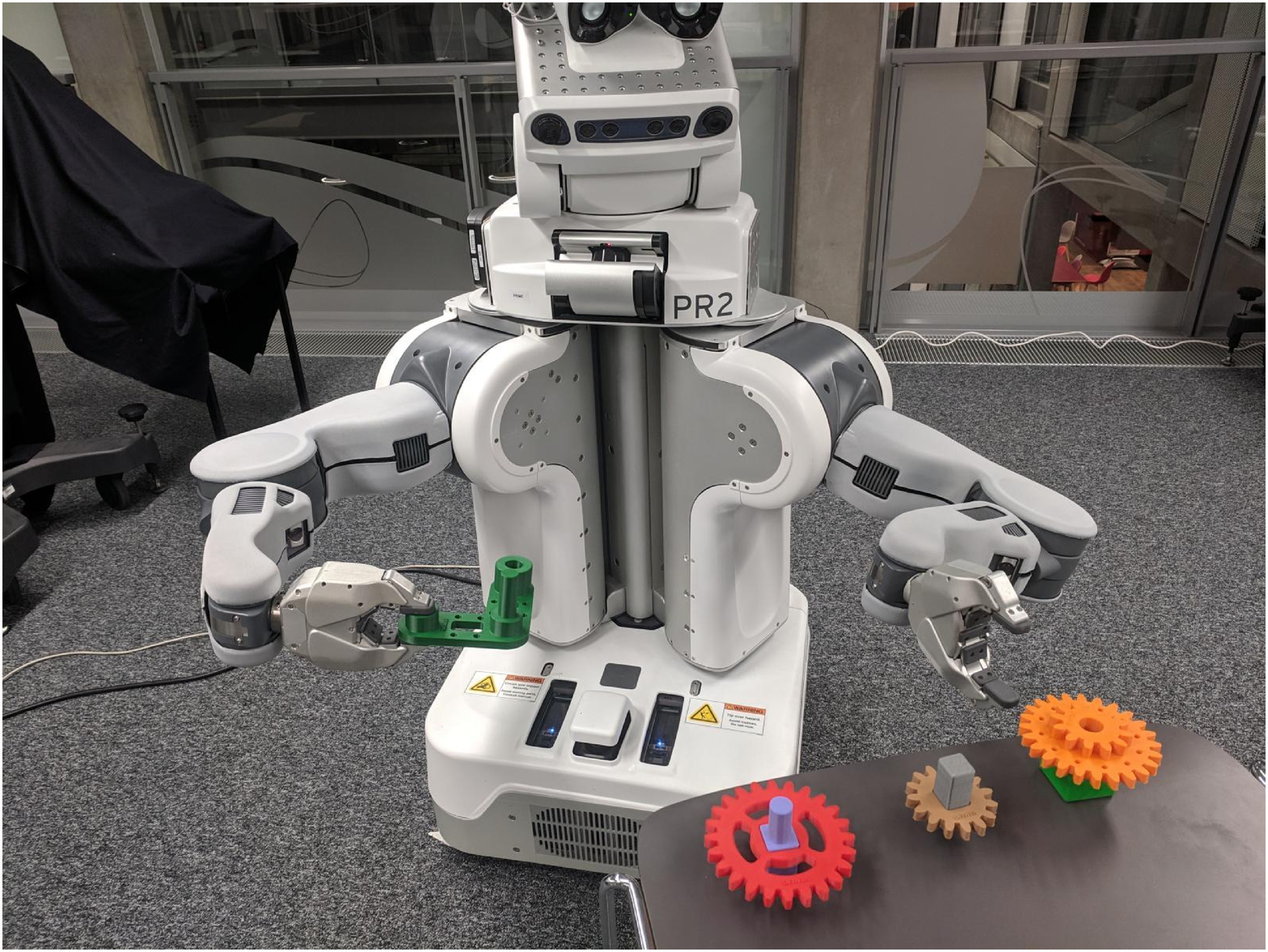}\label{fig:robot_assembly}}
\hspace{0.02\linewidth}
\subfigure[Option 4 of inserting a gear on a peg]{\includegraphics[width=0.73\linewidth]{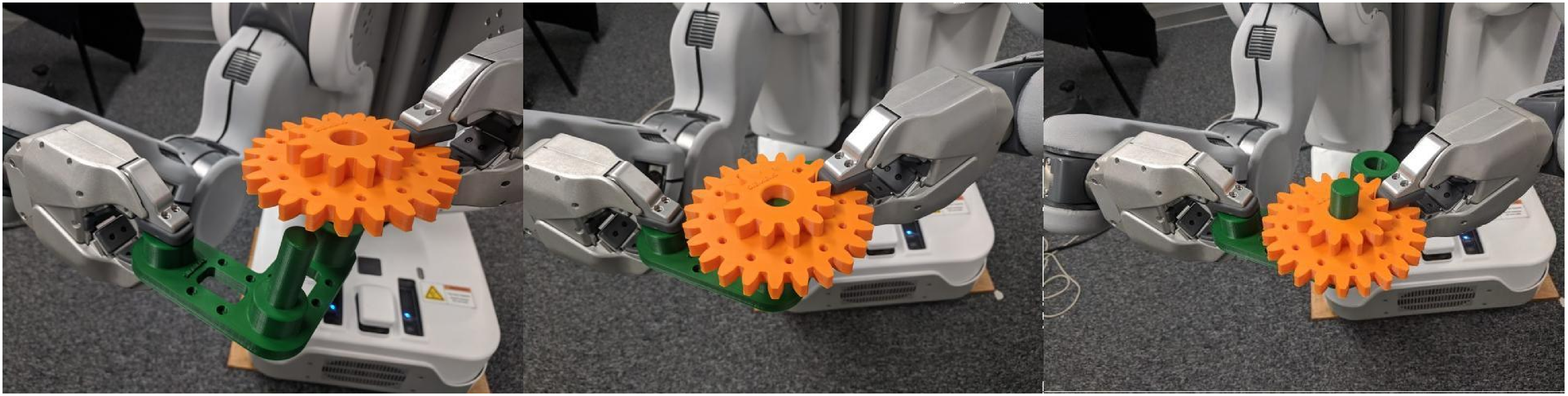}\label{fig:gear_peg}}
\caption{The gear assembly problem executed by the robot. The execution of option 4, (Section.\ref{sec:prob}) is shown on the right.}
\end{figure}

Modern Deep Reinforcement Learning (DRL) approaches focus on generating small policies that solve individual problems (pick up/grasp/push) \cite{klissarov2017learnings}, or longer range end to end solutions illustrated in modern games.
Typically, in order to provide a good initialization for the optimization algorithm, expert demonstrations are provided either through human demonstration \cite{Argall2009} or through the use of a motion planner as an initial approximation to the solution \cite{thomas2018learning}. In problems that allow for a simulator to be used as part of the inference and learning procedure, DNN \& tree based approaches have shown great promise in solving Chess, Go, Poker. To extend these methods to more general domains, a world dynamics model is required to approximate the environment as in \cite{ha2018world}. 

DynoPlan aims at extending the options framework in the following ways:
\begin{itemize}[noitemsep,nolistsep]
    \item We learn a dynamics model $s_{t+1} \sim \mathcal{D}(s_t, a_t)$ for each option that predicts the next state of the world given the current action; and 
    \item We learn a goal heuristic $\mathcal{G}(s_t)$ that gives a distribution as an estimate of how close the state is to completing the task, based on the demonstrations.
\end{itemize}
This allows for the higher level controller to perform reasoning about sequentially applying controllers in overlapping initiation sets for completing a task.

We aim to show that we can use off-the-shelf model-based controllers in parts of the state space, where their performance is already optimized, and model-free methods for states without correspondingly robust or easily scripted solutions, combining these two categories of controllers into a hybrid solution.

\section{Related Work}

Our method sits between learning policies over options as in \cite{barto2003recent}; and computing solutions using learning from demonstration such as through inverse reinforcement learning \cite{arora2018survey}. 
Reinforcement Learning is intrinsically based on the forward search of good states through experience. The update of the quality of an action at a particular state is performed by the iterative application of the Bellman equation. Performing updates in a model-free method must overcome the problems of sparse reward and credit assignment. Introducing a learned model that summarizes the dynamics of the problem can alleviate some scaling issues as in \cite{ha2018world}. However, searching for a general world model remains hard and we are not aware of methods that can achieve the desired performance levels in physical real world tasks. 
Such problems usually exhibit a hierarchical sequential structure - e.g. the \textit{waking up routine} is a sequence of actions, some of which are conditioned on the previous state of the system.

The options framework provides a formal way to work with hierarchically structured sequences of decisions made by a set of RL controllers. An option consists of a policy $\pi_\omega(a_t|s_t)$, an initiation set $\mathcal{I}$ and termination criteria $\beta_\omega(s_t)$ - probability of terminating the option or reaching the terminal state for the option. A policy over options $\pi_\Omega(\omega_t|s_t)$ is available to select the next option when the previous one terminates as shown by \cite{sutton1999between, precup2000eligibility}.

Temporal abstractions have been extensively researched by \cite{ iba1989heuristic, sutton1999between}. The hierarchical structure helps to simplify the control, allows an observer to disambiguate the state of the agent, and encapsulates a control policy and termination of the policy within a subset of the state space of the problem. This split in the state space allows us to verify the individual controller within the domain of operation  - \cite{andonov2015controller}, deliberate the cost of an option and increase the interpretability - \cite{harb2018waiting}.
Our method borrows this view of temporally abstracting trajectories and extends it by enforcing a dynamics model for each of the options allowing out agent to incorporate hindsight in its actions.

To expedite the learning process, we can provide example solution trajectories by demonstrating solutions to the problem. This can be used to learn safe policies \cite{huang2018learning}. Alternatively, it can be used to calculate the relative value of each state by Inverse Reinforcement Learning \cite{arora2018survey}. 
For instance, we can expect that agents would be approximately rational in achieving their goal, allowing \cite{baker2009action} to infer them. 
Exploring the space of options may force us to consider ones that are unsafe for the agent. \cite{brown2019risk} rephrases the active inverse reinforcement learning to optimize the agents policy in a risk-aware method. 
Our work partitions the space of operation of each option, allowing that area to inherit the safety constraints that come associated with the corresponding policy.

\section{Problem Definition}
\label{sec:def}

We assume there exists an already learned set of options $\mathcal{O}=\{o_1, o_2, ..., o_N\}$ and a set of tasks $\mathcal{K}= \{K_1, K_2, \dots, K_L\}$. Each option $o_\omega$ is independently defined by a policy $\pi_\omega(s) \rightarrow a$, $s \in \mathcal{S}_\omega$, $a \in \mathcal{A}_\omega$, an initiation set $\mathcal{I}_\omega, \mathcal{I}_\omega \subseteq \mathcal{S}_\omega$ where the policy can be started, and a termination criteria $\beta_\omega$. We extend the options formulation by introducing a forward dynamics model $s_{t+1} \sim \mathcal{D}_\omega(s_t)$, which is a stochastic mapping, and a goal metric $g \sim \mathcal{G}_{K_j}(s_t), 0 \leq g \leq 1$, that estimates the progress of the state $s_t$ with respect to the desired world configuration. We aim for $\mathcal{G}_{K_j}$ to change monotonically through the demonstrated trajectories. The state space of different options $\mathcal{S} = \{\mathcal{S}_1, \mathcal{S}_2, .., \mathcal{S}_N\}$ can be different, as long as there exists a direct or learnable mapping between $\mathcal{S}_i$ and $\mathcal{S}_j$ for some part of the space.

We aim to answer the question whether we can construct a hybrid hierarchical policy $\pi_\Omega(\omega_t |s_t)$ that can plan the next option $o_{\omega_t}$ that needs to be executed to bring the current state $s_t$ to some desired $s_{final}$ by using the forward dynamics model $\mathcal{D}_\omega$ in an $n$-step MPC look-ahead using a goal metric $\mathcal{G}_K$ that evaluates how close $s_{t+n}$ is to $s_{final}$.

\section{Method}

At a particular point $s_t$ when $o_\omega$ is active, we can compute how successful is following the policy given these conditions up to a particular time horizon. The action given by the policy is $a_t=\pi_\omega(s_t)$, and following the dynamics model we can write that $s_{t+1}=\mathcal{D}_\omega(s_t, a_t)=\mathcal{D}_\omega(s_t, \pi_\omega(s_t))$. As the dynamics model is conditioned on the policy, we can simplify the notation to $s_{t+1}=\mathcal{D}_\omega(s_t)$.
Chaining it for n steps in the future we obtain 
\( s_{t+n}=\mathcal{D}_\omega \circ \mathcal{D}_\omega  \circ \dots \circ \mathcal{D}_\omega(s_t) =\mathcal{D}^n_{\omega}(s_t) \).     
Thus, a policy over policies can sequentially optimize 
\vspace{-0.4em} \begin{align}
\pi_\Omega(\omega_t|s_t)=\argmax _\omega \left(\mathbb{E}\left[\1_{\mathcal{I}_\omega} (s_t) \cdot \mathcal{G} \circ \mathcal{D}^n_\omega (s_t) \right] \right)  
\end{align} 
After choosing and evaluating the optimal $\pi_\Omega$ with respect the above criterion, another controller can be selected until the goal is reached.


\section{Experimental Setup}
\label{sec:prob}
We perform two sets of experiments to showcase the capability of using the structured hierarchical policy by performing MPC future predictions at each step on a simulated MDP problem and on a gear assembly task on the PR2 robot.

\textbf{Simulated MDP}~~ In the first we use the standard 19-state random walk task as defined in \cite{harutyunyan2018learning} and seen on Figure.~\ref{fig:mdp_problem}. The goal of the agent is to reach past the $19^{th}$ state and obtain the $+1$ reward. The action space of the agent is to go ``left'' or ``right''. There also exist 5 options defined as in Section.~\ref{sec:def}, with the following policies: (1-3) policies that go ``right'' with a different termination probabilities $\beta = \{0.9, 0.5, 0.2\}$; (4) random action; (5) policy with action to go ``'left'' with $\beta=0.5$. We assume that there exists a noisy dynamics model $\mathcal{D}_\omega$ and the goal evaluation model $\mathcal{G}_{MDP}$, obtained from demonstrations, that have probability of mispredicting the current state or its value of $0.2$.

\textbf{Gear Assembly}~~ In this task the PR2 robot needs to assemble a part of the Siemens Challenge\footnote{ The challenge can be seen https://new.siemens.com/us/en/company/fairs-events/robot-learning.html }, which involves grasping a compound gear from a table, and placing it on a peg module held in the other hand of the robot. A human operator can come in proximity to the robot, interfering with the policy plan. We have received expert demonstrations of the task being performed, as well as access to a set of option that \textbf{(1)} picks the gear from the table; \textbf{(2)} quickly moves the left PR2 arm in proximity to the other arm; \textbf{(3)} cautiously navigates the left PR2 arm to the other avoiding proximity with humans; \textbf{(4)} inserts the gear on the peg module. Policy (2) relies exclusively on path planning techniques, (4) is fully neural networks learned and (1, 3) are a mixture between a neural recognition module recognizing termination criteria and motion planning for the policy. The options share a common state space of the robots' joint angles. The initiation set of all policies is $\mathcal{R}^{12}$. The terminal criteria $\beta$ for $o_{1,2}$ is inversely proportional to the closeness to a human to the robot; for $o_{2-4}$ is the proximity to a desired offset from the other robot hand.

The dynamics model for each option is independent and is represented either as part of the motion planner, or similarly to the goal estimator - a neural networks working on the joint angle states of both arms of the robot.

In cases of options with overlapping initiation sets (i.e. options 1, 2, 3 all work within $\mathcal{R}^{12}$), we can softly partition the space of expected operation by fitting a Gaussian Mixture Model $\mathcal{F}_M$ on the trajectories of the demonstrations, where $s, s \sim \mathcal{F}_M$ is a sample state from the trajectory. $\mathcal{F}$ is a set of $M$ Gaussian Mixtures $\mathcal{F} = \{N_i(\mu_i, \Sigma_i) |_{i=1..M}\}$, and $J_k$ is a subset of $\mathcal{F}_M$, where samples from $J_k$ correspond to samples from trajectory of option $k, 1 \le k \le N$.
We can thus assess the likelihood of a particular option working in a state $s \sim \pi_j$ by evaluating
\(
    \mathcal{L}(s| \pi_j, \mathcal{F}_M)= \max_p \left[p(s|\mu_i, \Sigma_i) \right]_{\mu_i, \Sigma_i \in J}~ (5),
    \label{eq:parition}
    \)
This gives us the safety region, in which we expect the policy to work. By using the overlap between these regions, we can move the state of the system in a way that reaches the desired demonstrated configuration.

\section{Results}

We aim to demonstrate the viability of using the options dynamics as a method for choosing a satisfactory policy. The dynamics can be learned independently of the task, and can be used to solve a downstream task.

\begin{figure}[ht]
\centering
\subfigure[MDP Problem]{\includegraphics[width=0.3\linewidth]{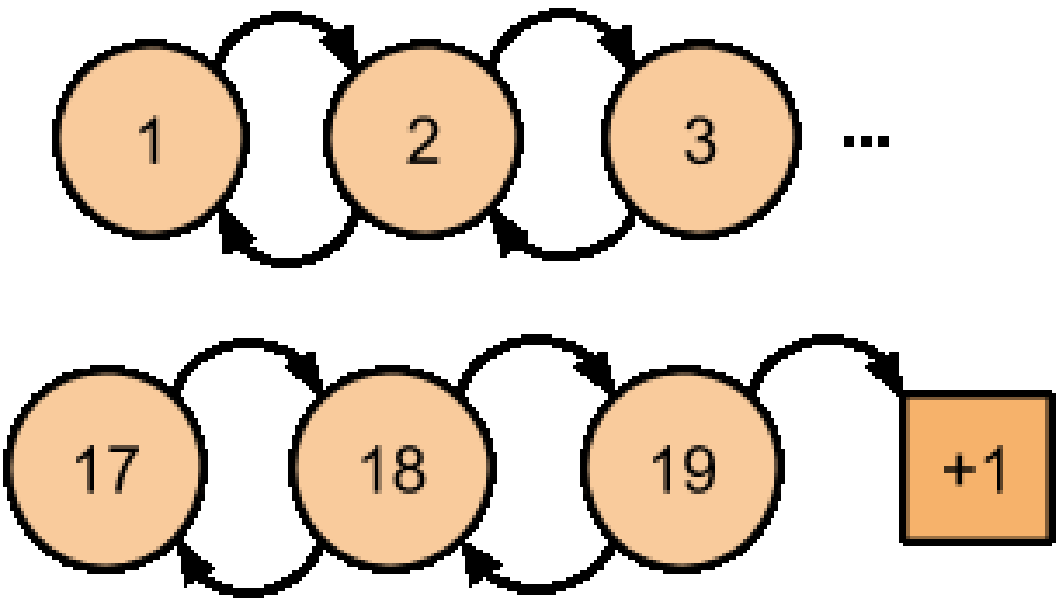}\label{fig:mdp_problem}}
\hspace{0.02\linewidth}
\subfigure[MDP Solution]{\includegraphics[width=0.47\linewidth]{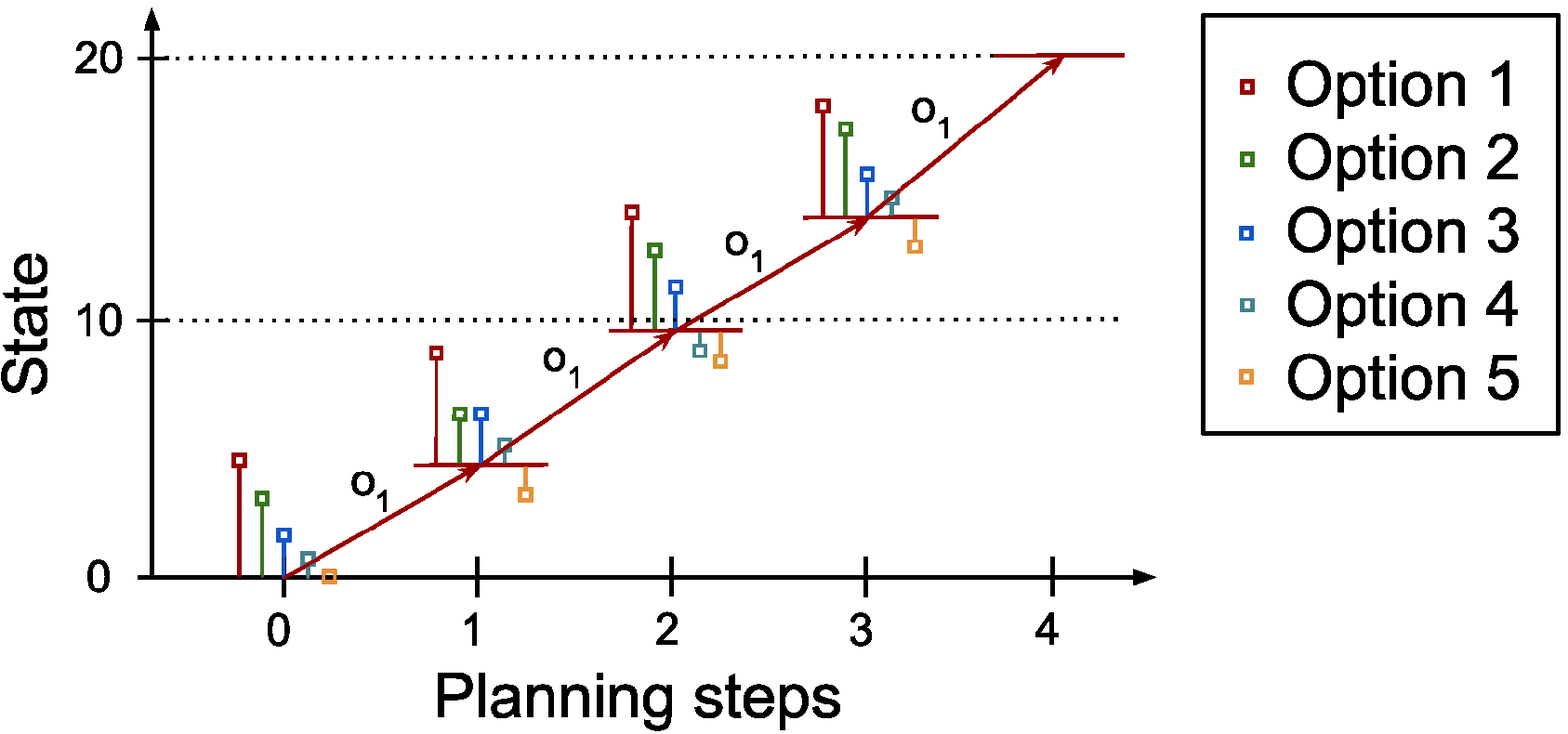}\label{fig:mdp_solution}}
\caption{(a) The 19-state MDP problem. The action space of the MDP is to move ``left'' or ``right''. The goal of the MDP problem is to reach past state 19 and obtain the +1 reward, which is equivalent to a termination state 20.
(b) MDP solution. At timestep 0, a rollout of the 5 options is performed with the dynamics model. The expected resulting state is marked as blue vertical bars. The best performing option is used within the environment to obtain the next state - the red line at state 5 and planning step 1. This process is iterated until a desired state is reached.}
\end{figure}

\textbf{Simulated MDP}~~ The target solution shows the feasibility and compares the possible solutions by using different options. In Figure.~\ref{fig:mdp_solution}, we can see that we reach the optimal state in just 4 planning steps, where each planning step is a rollout of an option. We can see the predicted state under the specified time horizon using different options. This naturally suggests the use of the policy $\pi_1$ that outperforms the alternatives ($\pi_1$ reaches state 6, $\pi_2$ - state 4, $\pi_2$ - state 3, $\pi_3$ - state 1, $\pi_4$ - state 1, $\pi_5$ - state 0). Even though the predicted state differs from the true rollout of the policy, it allows the hierarchical controller to use the one, which would progress the state the furthest. The execution of some options (i.e. option 5 in planning steps 1, 2, 3) reverts the state of the world to a less desirable one. By using the forward dynamics, we can avoid sampling these undesirable options.

\begin{figure}[ht]
\centering
\subfigure[Goal Heuristic]{\includegraphics[width=0.35\linewidth]{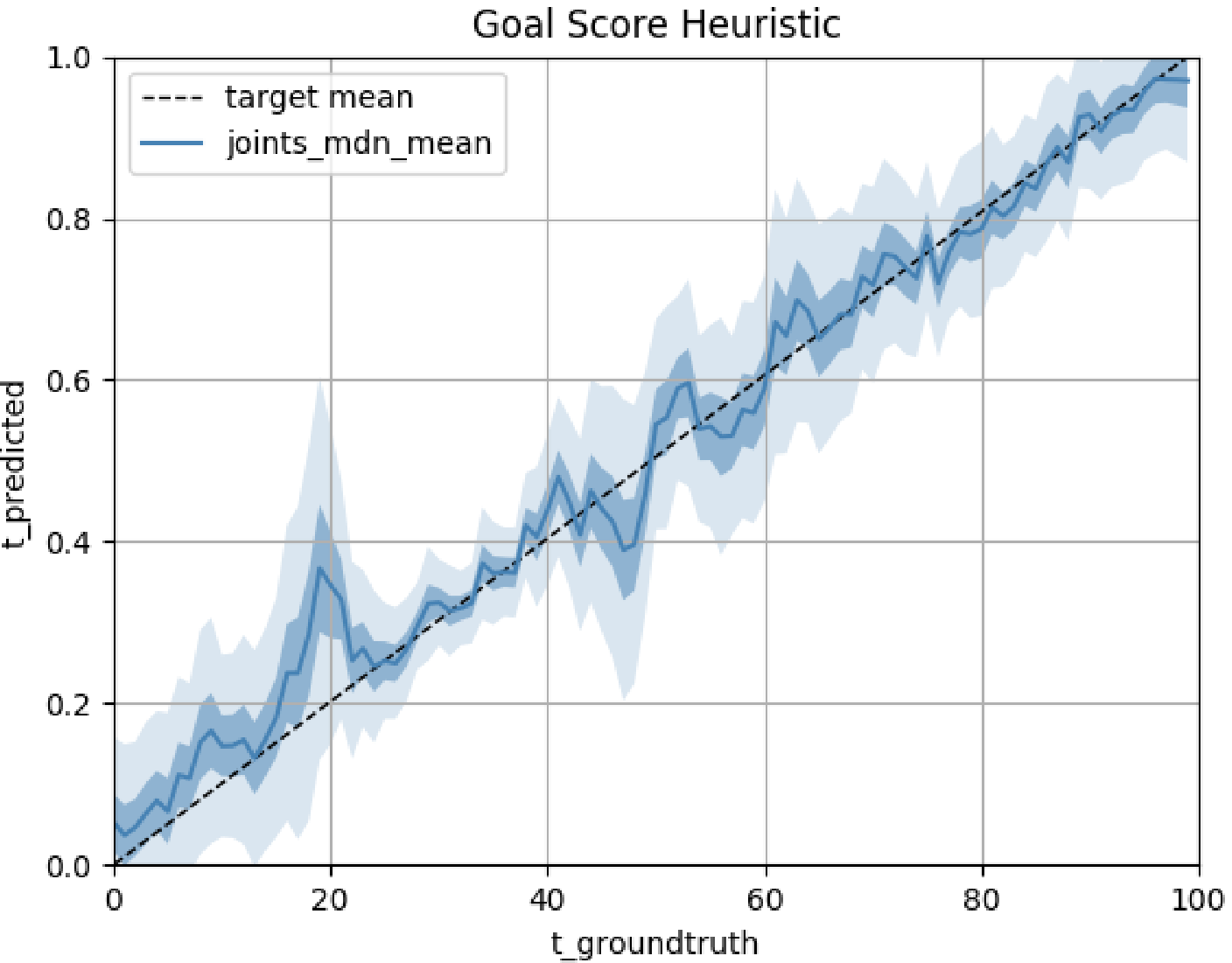} \label{fig:heuristics}}
\hspace{0.04\linewidth}
\subfigure[t-SNE visuazlization of the controllers states.]{\includegraphics[width=0.37\linewidth]{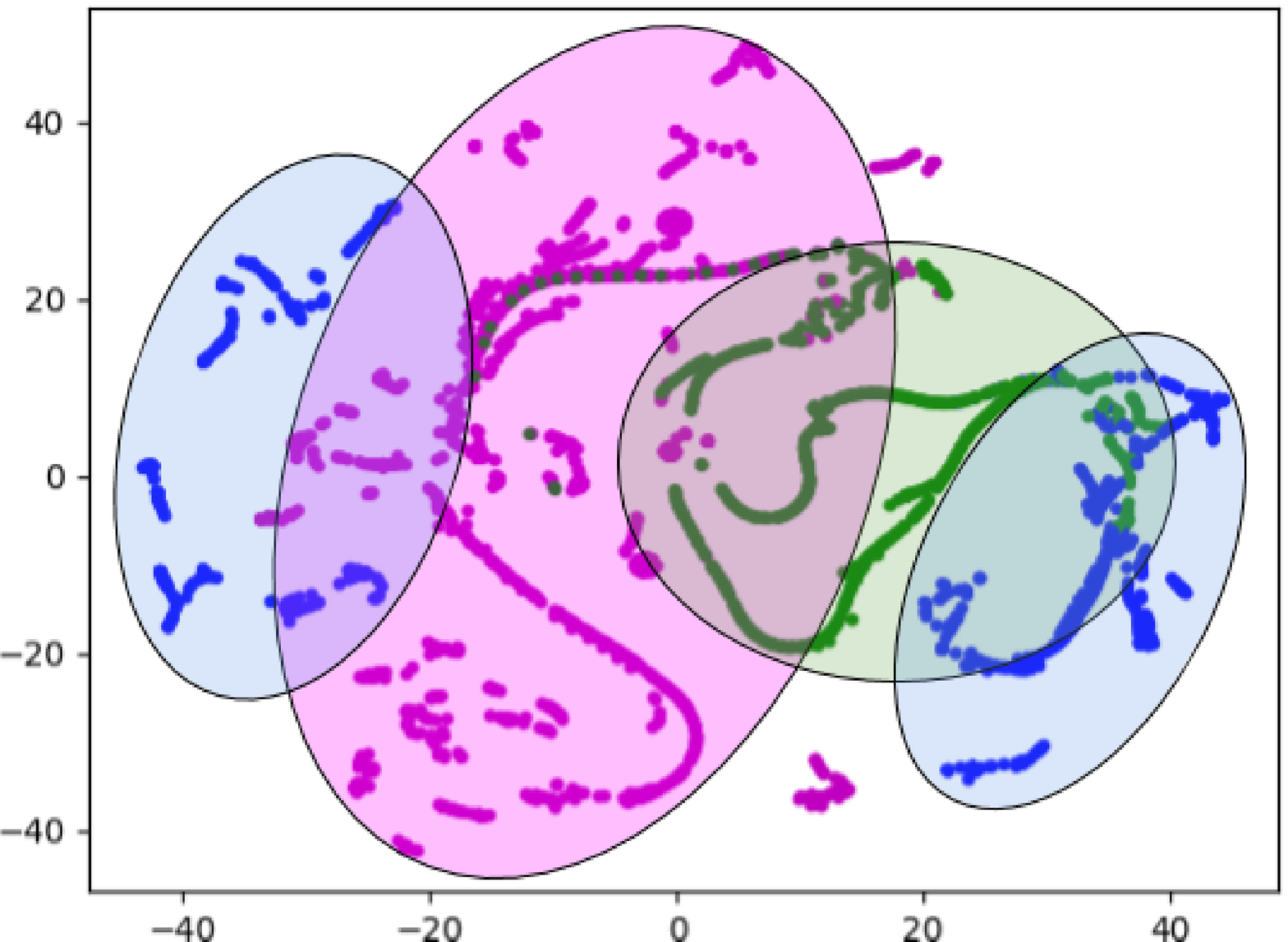}\label{fig:tsne}}
\caption{(a) The learned heuristics about how close the current state is to the demonstrated goal state.  (b) t-SNE plot of the controllers state during a set of trajectories. \textit{Magenta} - $o_1$ for grasping the object, \textit{Green} - $o_2$ and $o_3$ for navigating to the assembly with and without a human intervention and \textit{Blue} - $o_4$ for inserting the gear onto the peg. The shaded regions illustrate the regions of control for the different policies.}
\label{fig:gears_assembly_results}
\end{figure}

\textbf{Gear Assembly}~~ We obtained 10 demonstrations of the task being performed. In Figure.~\ref{fig:heuristics}, we show the performance of the goal estimator network on an independent trial. We can observe that the state goal metric estimator closely tracks the expected ground truth values along the trajectory. This provides reasonable feedback that can be used by $\pi_\Omega$ to choose an appropriate next policy.

Similarly, in Figure.~\ref{fig:tsne} we show the t-SNE of the trials of the robot trajectories that have no interruptions and some in which a human enters the scene and interferes with the motion of robot, forcing a change of policy to occur. We see that there is natural split in the states in which different options have been activated. We can notice that the overlap of the region of activation for the different policies allows the robot to grasp, navigate to, and insert the gear into the assembly by following these basins of the policies initiation. 
By following Eq.\ref{eq:parition} we can therefore create state space envelope of action of each option. The corresponding part of the state space, conditioned on the executed option, can have the safety constraints enforced by the underlying control method for the option.

\section{Conclusion}
We present DynoPlan - a hybrid hierarchical controller where by extending the options framework, we can rephrase the learning of a top level controller to an MPC planning solution. By unrolling the future states of each option, where each can be assessed on the contribution of furthering the agents intent based on the goal heuristic, we can choose the one best satisfying the problem requirements. This method of action selection allows to combine motion planning with neural network control policies in a single system, whilst retaining the completeness and performance guarantees of the work space of the associated options.

\bibliography{rldm}

\bibliographystyle{unsrt}
\renewcommand*{\bibfont}{\footnote}

\end{document}